\documentclass[sigconf,screen]{acmart}

\def\BibTeX{{\rm B\kern-.05em{\sc i\kern-.025em b}\kern-.08emT\kern-.1667em\lower.7ex\hbox{E}\kern-.125emX}}
    
\setcopyright{none} 
\renewcommand \footnotetextcopyrightpermission[1]{} 
\usepackage{graphicx} 
\usepackage{multirow}
\usepackage[ruled,linesnumbered]{algorithm2e}
\makeatletter
\let\oldnl\nl
\newcommand{\nonl}{\renewcommand{\nl}{\let\nl\oldnl}}
\makeatother
\begin{document}
%
\title{ST-UNet: A Spatio-Temporal U-Network for \\ Graph-structured Time Series Modeling}

%

%

\author{Bing Yu}
\authornote{Both authors contributed equally to this research.}
\email{byu@pku.edu.cn}
\affiliation{%
  \institution{School of Mathematical Sciences, Peking University}
  \city{Beijing}
  \country{China}
  \postcode{100871}
}

\author{Haoteng Yin}
\authornotemark[1]
\email{htyin@pku.edu.cn}
\affiliation{%
  \institution{Center for Data Science (AAIS), Peking University}
  \city{Beijing}
  \country{China}
  \postcode{100871}
}

\author{Zhanxing Zhu}
\email{zhanxing.zhu@pku.edu.cn}
\authornote{Corresponding author.}
\affiliation{
  \institution{Center for Data Science, \\Peking University}
  \institution{Beijing Institute of Big Data Research}
  \city{Beijing}
  \country{China}
  \postcode{100871}
}

%
\begin{abstract}
The spatio-temporal graph learning is becoming an increasingly important object of graph study. Many application domains involve highly dynamic graphs where temporal information is crucial, e.g. traffic networks and financial transaction graphs. Despite the constant progress made on learning structured data, there is still a lack of effective means to extract dynamic complex features from spatio-temporal structures. Particularly, conventional models such as convolutional networks or recurrent neural networks are incapable of revealing the temporal patterns in short or long terms and exploring the spatial properties in local or global scope from spatio-temporal graphs simultaneously. To tackle this problem, we design a novel multi-scale architecture, Spatio-Temporal U-Net (ST-UNet), for graph-structured time series modeling. In this U-shaped network, a paired sampling operation is proposed in spacetime domain accordingly: the pooling (ST-Pool) coarsens the input graph in spatial from its deterministic partition while abstracts multi-resolution temporal dependencies through dilated recurrent skip connections; based on previous settings in the downsampling, the unpooling (ST-Unpool) restores the original structure of spatio-temporal graphs and resumes regular intervals within graph sequences. Experiments on spatio-temporal prediction tasks demonstrate that our model effectively captures comprehensive features in multiple scales and achieves substantial improvements over mainstream methods on several real-world datasets.
\end{abstract}

%
%
%
%

%

%
\maketitle

\section{Introduction}
With the latest success of extending deep learning approaches from regular grids to structured data, graph representation learning has become an active research area nowadays. Many real-world data such as social relations, biological molecules and sensor networks are naturally with a graph form. Recently, there has been a surge of interests in exploring and analyzing the representation of graphs for tasks like node classification and link prediction \cite{kipf2016semi,hamilton2017inductive,gao2018large}. However, among those studies, the dynamic graph has received relatively less attention than the static graph that consists of fixed node values or labels. The spatio-temporal graph is one of typical dynamic graphs, with varying input for each node along time axis, e.g. traffic sensor streaming and human action sequences. In this work, we systematically study the dynamic graph in spacetime domain, with an aim to develop a principled and effective method to interpret the spatio-temporal graph and to forecast future values or labels of certain nodes thereof, or to predict the whole graph in the next few time steps.

In the field of spatio-temporal data, videos are a well-studied example, whose successive frames consistently share spatial and temporal structures. By leveraging different types of neural networks, a hybrid framework is constructed to exploit such spatio-temporal regularities within video frames, for instance, applications in weather radar echoes \cite{xingjian2015convolutional} and in traffic heatmaps \cite{zhang2018predicting}. In this case, each frame in the video firstly passes through convolution neural networks (CNN) for visual feature extraction, and then followed by recurrent neural networks (RNN) for sequence learning. Even though images can be regarded as special cases of graphs, widely used deep learning models still face significant challenges in applying to spatio-temporal graphs. First, graph-structured data are generated from non-Euclidean domain, which may not align in regular grids as required by existing models. Second, compared to grid-like data, there is no spatial locality or order information among nodes of a graph. Due to such irregularities, standard operations (for example, convolution and pooling) are not directly applicable to graph domain.

To bridge the above gap, \cite{bruna2013spectral} proposes graph convolutional networks (GCNs) redefining the notion of the convolution and generalizing it to arbitrary graphs based on spectral graph theory. The introduction of GCNs boosts the latest rapid development of graph study. Moreover, it has been successfully adopted in a variety of applications where the dynamic graph is strongly associated. For instance, in action recognition, human action sequences can be assembled as a spatio-temporal graph, where body joints are constituted as a series of skeleton graph changing along time axis. Correspondingly, \cite{AAAI1817103} designs a GCN-based model to capture the spatial patterns of skeleton sequences as well as the temporal dynamics contained therein. In traffic forecasting, each sensor station streams the traffic status of a certain road within a traffic network. In this sensor graph, the spatial edges are weighted by the pair-wised distance between stations in the network while the temporal ones are connected by the same sensors between adjacent time frames. Recent studies have investigated the feasibility of combining GCNs with RNN \cite{li2018dcrnn_traffic} or CNN \cite{yu2018spatio} for traffic prediction on road networks. GCN-based models obtain considerable improvements compared to traditional ones that typically ignore the spatio-temporal correlations and lack in the capability for handling structured sequences.

In order to accurately understand local and global properties of dynamic graphs, it is necessary to process the data through multiple scales. The spatio-temporal graph particularly requires such scale-spanning analysis since its particularity and complexity in spacetime domain. However, most mainstream methods have overlooked such principle, partially because of the difficulties of extending existing operations like the pooling to graph data. Nevertheless, multi-scale modeling of the dynamic graph has the similarity with the pixel-wise prediction task, as an image pixel corresponding to a graph node. U-shaped networks with U-Net \cite{ronneberger2015u} as the representative achieve state-of-the-art performance on pixel-level prediction, whose architecture has high representational capacity of both the local distributed and the global hidden information within the input. Thus, it is particularly appealing to apply such U-shaped design to modeling dynamic graphs.

In this paper, we propose a novel multi-scale framework, Spatio-Temporal U-Net (ST-UNet), to model and predict graph-structured time series. To precisely capture the spatio-temporal correlations in dynamic graphs, we firstly generalize the U-shaped architecture from images to spatio-temporal graphs. ST-UNet employs multi-granularity graph convolution for extracting both generalized and localized spatial features, and adds dilated recurrent skip-connections for capturing multi-resolution temporal dependencies. Under the settings of ST-UNet, we define two essential operations of the framework accordingly: the spatio-temporal pooling (ST-Pool) operation samples nodes to form a smaller graph from the output of deterministic graph partition \cite{maue2007engineering} and abstracts time series at multiple temporal resolutions through skip connections between recurrent units. Consequently, the unpooling (ST-Unpool) as a paired operation restores the original structure and temporal dependency of dynamic graphs based on previous settings in the downsampling. To better localize the representation from the input, higher-level features retrieved from the pooling part are concatenated with the upsampled output. Overall, with contributions of hierarchical U-shaped design, ST-UNet is able to effectively derive multi-scale features and precisely learn representations from the spatio-temporal graph.

\section{Related Work}
Following spectral-based formulation \cite{bruna2013spectral,niepert2016learning,defferrard2016convolutional}, the graph convolution operator `$*_G$' is introduced as the multiplication of a graph signal $x \in \mathbb{R}^n$ with a kernel $g_{\theta} (\Lambda) = diag(\theta)$, where $\theta$ is a vector of Fourier coefficients, as
\begin{equation}
\label{eq:gconv}
g_{\theta} *_G x = g_{\theta} (L) x = g_{\theta} (U \Lambda U^{T}) x = U g_{\theta} (\Lambda) U^{T} x
\end{equation}
where $U \in \mathbb{R}^{n \times n}$ is the graph Fourier basis, which is a matrix of eigenvectors of the normalized graph Laplacian $L = I_n - D^{-\frac{1}{2}} W D^{-\frac{1}{2}}=U \Lambda U^{T} \in \mathbb{R}^{n \times n}$ ($I_n$ is an identity matrix and $D \in \mathbb{R}^{n \times n}$ is the diagonal degree matrix of adjacency matrix $W$ with $D_{i i}=\Sigma_{j} W_{i j}$); while $\Lambda \in \mathbb{R}^{n \times n}$ is the diagonal matrix of eigenvalues of $L$ \cite{shuman2012emerging}. In order to localize the filter, the kernel $g_{\theta}$ can be restricted to a truncated expansion of Chebyshev polynomials $T_k(\cdot)$ to $K-1$ order with the rescaled $\tilde{\Lambda} = 2\Lambda / \lambda_{\max} - I_{n}$ as $g_{\theta} (\Lambda) = \sum_{k=0}^{K-1} \theta_{k} T_{k} (\tilde{\Lambda})$, where $\theta_{(\cdot)} \in \mathbb{R}^{K}$ is a vector of Chebyshev coefficients \cite{hammond2011wavelets}. Hence, the graph convolution can then be expressed as,
\begin{equation}
\label{eq:cheb}
g_{\theta} *_G x = g_{\theta} (L) x = \sum_{k=0}^{K-1} \theta_{k} T_{k} (\tilde{L}) x,
\end{equation}
where $T_{k} (\tilde{L}) \in \mathbb{R}^{n \times n}$ is the Chebyshev polynomial of order $k$ evaluated at the rescaled Laplacian $\tilde{L} = 2L / \lambda_{\max} - I_{n}$.

Apart from convolutional operations on graphs, there are also several recent studies focusing on structured sequence learning. Structured RNN \cite{jain2016structural} attempts to fit the spatio-temporal graph into a mixture of recurrent neural networks by associating each node and edge to a certain type of the networks. Based on the framework of convLSTM \cite{xingjian2015convolutional}, graph convolutional recurrent network (GCRN) \cite{seo2018structured} is firstly proposed modeling structured sequences by replacing regular 2D convolution with spectral-based graph convolution. And it has set a trend of GCN-embedded designs for the follow-up studies \cite{li2018dcrnn_traffic,yu2018spatio}.
The fast pooling operation on graph signals was firstly introduced by \cite{defferrard2016convolutional}. Recently, graph U-Net \cite{gao2019graph} brings pooling and upsampling operations to graph data. However, the applicable scope of these operations  is bounded by the static graph. Especially, graph U-Net introduces extra training parameters for node selection during the pooling procedure. Furthermore, its pooling operation does not keep the original structure of the input graph that may raise an issue for those tasks whose local spatial relations are critical.

\section{Methodology}
In this section, we start with the definition of the spatio-temporal graph and the problem formulation of prediction tasks on it. The special design of U-shaped network is elaborated in the following with essential operations of pooling and upsampling defined on the spatio-temporal graph. Base on the above advances, a multi-scale architecture, Spatio-Temporal U-Net, is introduced for graph-structured time series modeling eventually.

\begin{figure*}
  \centering
  \includegraphics[width=0.75\textwidth]{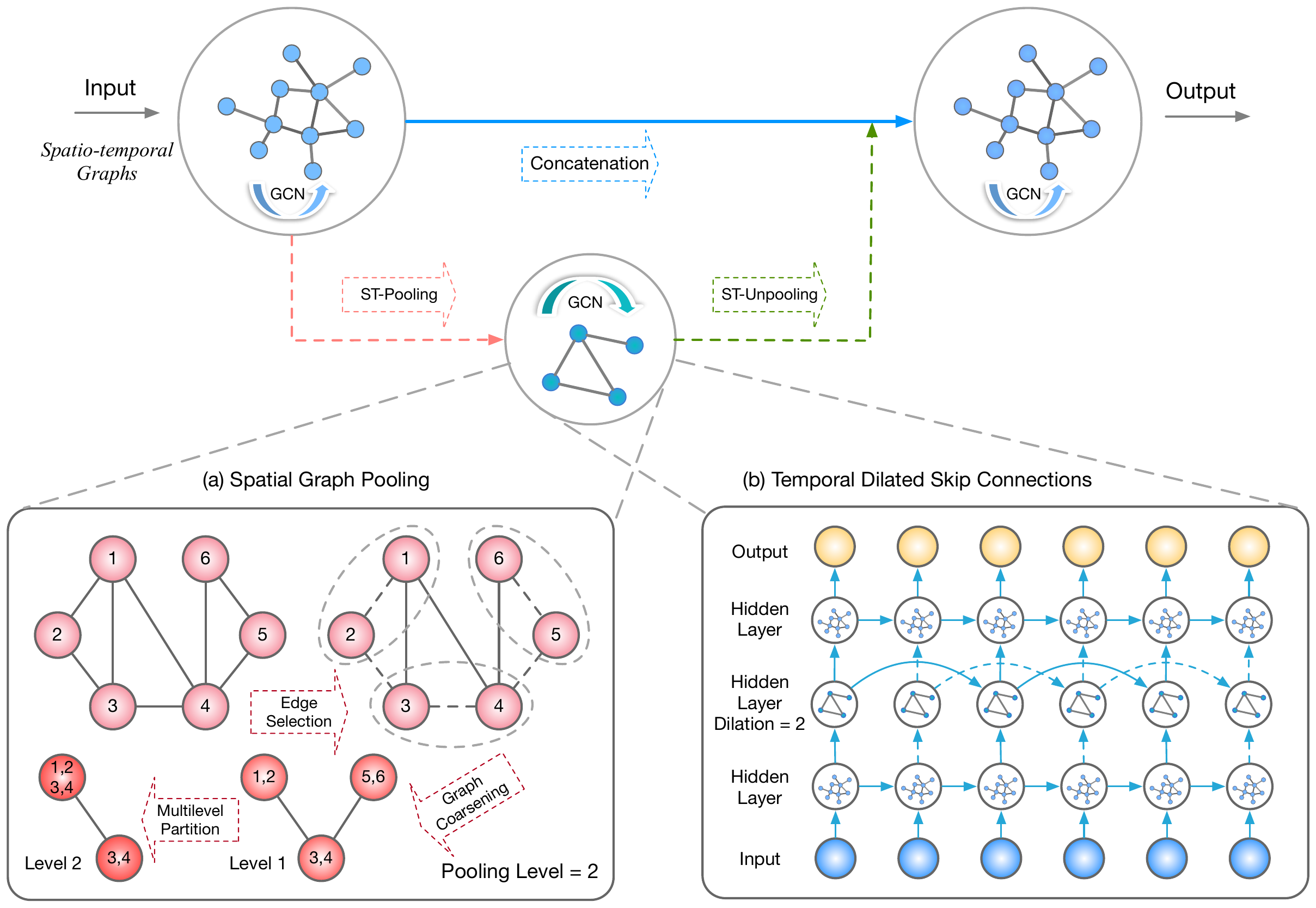}
  \caption{\label{fig:model}An illustration of the proposed Spatio-Temporal U-Net architecture. ST-UNet employs graph convolutional gated recurrent units (GCGRU) as its backbone. In this example, the proposed framework contains three GCGRU layers formed as a U-shaped structure with one ST-Pool and one ST-Unpool applied in one side respectively. Spatio-temporal features obtained from the input are downsampled into multi-resolution representations through a ST-Pooling operation. As subgraph (a) represents, the input graph at each time step is equally coarsened into nearly a quarter of its original size at the level 2 combining with feature pooling regarding the channel dimension. Meanwhile, the temporal dependency of the input sequence is dilated to 2 with skip-connections crossing every other recurrent unit, as shown in subgraph (b). The ST-Unpooling, as a reverse operation, restores the spatio-temporal graph into its original structure with upsampling in spatial features and resumes regular dependencies of time series concurrently. To assemble a more precise output with better localized representations, high-level features of the pooling side are fused with the upsampled output through a skip connection at the same level. The final output can be utilized for predicting node attributes or the entire graph in the next few time steps.}
\end{figure*}

\subsection{Spatio-temporal Graph Modeling}
Suppose spatio-temporal data are gathered through a structured spatial region consisting of $N$ nodes. Inside each node, there are $D$ measurements which vary over time. Thus, observation at any time can be represented by a feature vector $x \in \mathbb{R}^{D}$. Moreover, data collected over the whole region are able to be expressed in terms of a feature matrix $X = \{x_1, x_2,~\ldots~,x_N\} \in \mathbb{R}^{N \times D}$. As time goes by, a chronological sequence of matrices $X_1, X_2,~\ldots,~X_t$ is accumulated, which can be further formalized as the spatio-temporal graph defined as follows.

\begin{definition} [Spatio-temporal Graph]
A spatio-temporal graph is an attributed graph with a time-variable feature matrix $\mathcal{X}$. It is defined as $\mathcal{G} = (\mathcal{V}, \mathcal{E}, W, \mathcal{X})$ where $\mathcal{V}$ is the set of $|\mathcal{V}|=N$ vertices, $\mathcal{E}$ is the set of edges, and $W \in \mathbb{R}^{N \times N}$ is an adjacency matrix recording the weighted connectedness between two vertices. Contrary to the static graph, node attributes of the spatio-temporal one evolve over time as $\mathcal{X} = \{X_1, X_2,~\ldots,~X_T\} \in \mathbb{R}^{T \times N \times D}$, where $T$ is the length of time steps and $D$ is the dimension of features in each node.
\end{definition}

In practice, due to structural properties of the data, spatio-temporal graph modeling can be formulated as the prediction task of graph-structured time series. The objective of this task is to accurately predict future attributes of nodes in a given spatio-temporal graph based on historical records, which is formally described below.

\begin{definition} [Spatio-temporal Prediction]
Spatio-temporal prediction aims to forecast the most likely future length-$H$ sequence of node attributes in a graph given the previous $J$ observations:
\begin{equation}
\begin{aligned}
& \hat {X}_{t + 1},~\ldots,~\hat {X}_{t + H} = \\
& \underset {X_{t + 1},~\ldots,~X_{t + H}} {\arg \max} P \left(X_{t + 1},~\ldots,~X_{t + H} | X_{t - J + 1},~\ldots,~X_{t}; \mathcal{G} \right)
\end{aligned}
\end{equation}
where $X_t \in \mathcal{X} $ is an observation of node attributes linked by a weighted graph $\mathcal{G}$ at time step $t$.
\end{definition}

\subsection{Pooling Operation on Spatio-temporal Graphs}
The spatio-temporal graph can be decomposed into two domains: graph-structured data in spatial while time series in temporal. As a result, it inherits the characteristics of structural complexity from graphs and dynamic complexity from sequences. Therefore, we discuss the downsampling approaches applied from the spatial and the temporal perspective respectively in this section. Lastly, a unified pooling operation is defined in spacetime domain.

\paragraph{Spatial Graph Pooling}
Pooling layers play a vital role in CNNs since its function of achieving feature reduction. It generally follows the convolutional layer to progressively reduce spatial resolution of feature maps and enlarge receptive fields, thereby controlling parameter overfitting and achieving better generalization. However, the standard pooling operation is not directly applicable to graph-structured data, since it requires distinct neighborhoods which are obviously not accessible from graphs. Besides the local pooling, there are operations imposed on the input generally that could bypass the requirement of locality information, such as global pooling and $k$-max pooling. But these pooling approaches also bring issues of limited flexibility and inconsistent selection \cite{gao2019graph}. 

It is indispensable for the pooling in extracting multilevel abstraction of graphs. Similar to the pooling operation used in \cite{defferrard2016convolutional}, we use the global path growing algorithm (GPA) \cite{maue2007engineering} instead to perform graph partitions by solving the maximum weight matching problem (noted as `MaxWeightMatching'). GPA is consist of the greedy algorithm and the improved path growing algorithm (PGA') \cite{drake2003linear}. Given a graph $G_t = (V, E, W)$ with $N$ nodes at time step $t$, PGA' finds an approximate solution to the problem with a subset of edges $M \subseteq E$ satisfying: 1) there are no two members of $M$ sharing an endpoint; 2) its total weights are the largest. Subsequently, it generates the partition through gradually removing edges in $M$ and merging nodes connected thereof, as the \textbf{Algorithm \ref{alg:pga}} describes. At each level, it reduces the size of a graph by the factor of two, producing a coarser graph corresponding to observing the data domain at a different resolution:
\begin{equation}
G^{\prime}_t = gPartition(G_t, p), G_t \in \mathcal{G}\\
\end{equation}
where $G^{\prime}_t = (V^{\prime}, E^{\prime}, W^{\prime})$ is a partitioned graph with $N^{\prime} \approx N/2^{p}$ nodes at the level $p$ which controls reduction scale of the input. $V^{\prime}$ is a set of super nodes, each element $v^{\prime}$ of which contains a disjoint subset of $V$. We use $f(\cdot)$ to denote mapping relations between nodes in $V$ and $V^{\prime}$. Formally, after the graph convolutional layer, we can acquire the convolved feature matrix of a coarser graph $G^{\prime}_t$ through the graph partition algorithm as
\begin{equation}
X_t^{\prime} = f(g_{\theta} *_G X_t),
\end{equation}
where $X_t \in \mathbb{R}^{N \times D}$ is a graph signal matrix with $D$ attributes in each node of $G_t$ while $X_t^{\prime} = \lbrace x_1^{\prime}, x_2^{\prime},~\ldots,~x_k^{\prime} | x_k^{\prime} \in \mathbb{R}^{P \times C} \rbrace$ is a length-$N^{\prime}$ feature matrix with $C$ channels in each node of $G^{\prime}_t$ and $P$ is the number of nodes contained in each super node $v^{\prime} \in V^{\prime}$. Finally, we employ the maximum or mean feature activation over $P$ nodes in partitioned regions to obtain pooled features in each $v^{\prime}$ of $G^{\prime}_t$ regarding the channel dimension as
\begin{equation}
\widetilde{X_t} = \operatorname{gPooling} (X_t^{\prime}),
\end{equation}
where $\widetilde{X_t} \in \mathbb{R}^{N^{\prime} \times C}$ is the output of spatial graph pooling with $C$-channel features on $N^{\prime}$ nodes. Figure \ref{fig:model} (a) shows an example of the spatial graph pooling. Since graph partition is calculated in advance, it makes the operation very efficient without introducing extra training parameters. Moreover, the scope of spatial graph pooling can be adjusted through the level $p$ which offers a precise control. In order to address the inconsistency issue in node selection, the deterministic result of graph partition in PGA' is equally applied to the graph $G_t$ at each time step.
\begin{algorithm}
\caption{\label{alg:pga} Graph Partition Algorithm (gPartition)} 
\KwIn{graph $\boldsymbol {G} = ( \boldsymbol {V} , \boldsymbol {E} ) , \boldsymbol {W}: \boldsymbol {E} \to \mathbb{R} ^ { \geq 0 }$; pooling level $\boldsymbol {p} $}

\KwOut{graph $\boldsymbol {G^{\prime}} = ( \boldsymbol {V^{\prime}} , \boldsymbol {E^{\prime}})$ with adjusted $\boldsymbol {W^{\prime}}$}

\nonl I. Edge Selection\\
A subset of edges, $M : = \emptyset$\;
\While{$E \neq \emptyset$}{
$Q : = \langle \rangle$\;
deterministically choose $v \in V$ with $\operatorname { deg } ( v ) > 0$\;
\While{$\operatorname { deg } ( v ) > 0$}{
let $e = ( v , u )$  be the heaviest edge adjacent to $v$\;
append $e$ to $Q$\;
remove $v$ and its adjacent edges from $G$\;
$v : = u$\;}
$M : = M \cup \text { MaxWeightMatching } ( Q )$\;
extend $M$ to a maximal matching\;}
\Return $M \subseteq E$\\
\nonl II. Graph Coarsening\\
\While{$M \neq \emptyset$}{
remove $e$ selected in $M$ from $G$\;
merge $v$ connected by $e$ as a super node $v^{\prime} \in V^{\prime}$\;
adjust adjacent edges $e^{\prime}$ with related weights $w^{\prime}$ of $v^{\prime}$\;
remove $e$ from $M$\;}
\Return $G^{\prime} $ \label{alg:loop}\\
\nonl III. Multilevel Partition\\
\While{$p > 1$}{
repeat Part I \& II with $G^{\prime}$ at step \ref{alg:loop} as input\;
$p : = p - 1$\;}
\end{algorithm}

\paragraph{Temporal Downsampling}
Recurrent neural networks and its variants have shown impressive stability and capability of tackling sequence learning problems. Conventional recurrent models such as long short-term memory (LSTM) \cite{hochreiter1997long} and gated recurrent units (GRU) \cite{chung2014empirical} are initially designed for regular sequences with fixed time intervals, which significantly limits their capacity for capturing complex data dependencies. Recently, several studies have explored how to expand the scope of recurrent units in RNNs to more sophisticated data like the spatio-temporal one. Based on fully-connected LSTM (FC-LSTM), \cite{xingjian2015convolutional} develops a modified recurrent network with embedded convolutional layers (convLSTM) to forecast spatio-temporal sequences. Inside each recurrent unit, convolutional operations with kernels are substituted for multiplications by dense matrices, which enables the network for handling image sequences. Afterwards, \cite{seo2018structured} extends this approach by replacing the standard convolution by the graph convolution for structured sequence modeling. Following the similar scheme, we leverage the GRU model and GCN layers as Graph Convolutional Gated Recurrent Units (GCGRU) to discover temporal patterns from graph-structured time series:
\begin{equation}
\begin{aligned}
\label{eq:gcgru}
z _ { t } &= \sigma \left( W_z *_G X _ { t } + U_z *_G h _ { t - 1 } \right)\\
r _ { t } &= \sigma \left( W_r *_G X _ { t } + U_r *_G h _ { t - 1 } \right)\\
h _ { t } ^ { \prime } &= \tanh \left( W_h *_G X _ { t } + U_h *_G (r _ { t } \odot h _ { t - 1 }) \right)\\
h _ { t } &= z _ { t } \odot h _ { t - 1 } + \left( 1 - z _ { t } \right) \odot h _ { t } ^ { \prime }
\end{aligned}
\end{equation}
where $\odot$ is the Hadamard product and $\sigma$ stands for non-linear activation functions. In this setting, $z_t$ and $r_t$ represent the gate of update and reset at time step $t$; while $h_t^{\prime}$ and $h_t$ denote the current memory content and final memory at current time step respectively. Both $W_{(\cdot)} \in \mathbb{R}^{K \times D_h \times D_x}$ and $U_{(\cdot)} \in \mathbb{R}^{K \times D_h \times D_h}$ are parameters of the size-$K$ graph convolutional kernel. We use the notion `$W_{(\cdot)} *_G X_t$' to describe the graph convolution between the graph signal $X_t$ and $D_h \times D_x$ filters which are the functions of the graph Laplacian $L$ parameterized by $K$-localized Chebyshev coefficients as Eq. \eqref{eq:cheb} notes. By stacking several graph convolutional recurrent layers, the adopted backbone GCGRU can be used as a seq2seq model for graph-structured sequence learning.

The above architecture may be enough to model structured sequences by exploiting local stationarity and spatio-temporal correlations. But it still suffers from the restriction of interpreting temporal dynamic through determinate periods. In terms of multi-timescale modeling, many attempts have been made to extend recurrent networks to various time scope, including phased LSTM \cite{neil2016phased} and clockwork RNNs \cite{koutnik2014clockwork}. Inspired by jumping design between recurrent units in \cite{chang2017dilated}, we insert the skip connection between gated recurrent units to learn graph-structured sequences in multilevel temporal dependencies. It also generates a dilation between successive cells, which is equivalent to abstract temporal features over a different resolution. Denote $c_t^l$ as the GCGRU cell in layer $l$ at time $t$. The dilated skip connection can be expressed as
\begin{equation}
c _ { t } ^ {l} = g \left( X _ { t } ^ {l} , c _ { {t - s}^l } ^ {l} \right),
\end{equation}
where $X_t^{l}$ is the input to layer $l$ at time $t$; $s^{l}$ denotes the skip length, also referred to the dilation of layer $l$; and $g(\cdot)$ represents the GRU cell and output operations. Figure \ref{fig:model} (b) provides a diagram of the proposed temporal downsampling implemented by the dilated recurrent skip-connections. Such hierarchical design of dilation brings in multiple temporal scales for recurrent units at different layers. It also contributes to broadening the range of temporal dependency as the regular jump connection does but with fewer parameters and high efficiency.

In summary, based on the proposals made in pooling on spatio-temporal data, we define \emph{spatio-temporal pooling} (ST-Pool) as the operation performing downsampling on a spatio-temporal graph by aggregating convolved features over non-overlapped partitions regarding the channel dimension on its spatial projection while dilating dynamic dependencies over recurrent units aligned in the same layer on its temporal projection.

\subsection{\label{sec:un}Spatio-temporal Unpooling Operation}
As the inverse operation of downsampling, the unpooling is crucial in the U-shaped network for recovering pooled features to their original resolution through upsampling. There are several approaches defined on grid-like data that could satisfy this aim, such as transposed convolution \cite{zeiler2011adaptive} and unpooling layers \cite{zeiler2014visualizing}. However, these operations are not directly applicable to spatio-temporal domain due to specialty and compositionality of its data. To this end, we propose \emph{spatio-temporal unpooling} (ST-Unpool) accordingly: to restore primary structure of the input, the operation utilizes the reversed mapping $f^{\prime}(\cdot)$ to place back merged nodes and edges from $G^{\prime}_t$ to $G_t$; to resume regular temporal dependencies between recurrent units, the output of each time step in a skip-connected layer are fed into a vanilla recurrent layer without further temporal dilation.

Meanwhile, we provide three strategies for upsampling node attributes from a coarser graph, namely, \emph{direct copy}, \emph{ordered deconv} and \emph{weighted deconv}. As the name suggests, the first approach directly copies features of a super node to each node it contains; while ordered deconvolution assigns parameterized features to each merged node based on its degree order. On top of ordered deconvolution, the weighted one concatenates structural information of merged nodes in a sub-graph as an embedded feature vector to upsampled features. All three methods of upsampling have been tested and compared in Section \ref{sec:up}.

\subsection{\label{sec:arch}Architecture of Spatio-Temporal U-Net}
Based on spatio-temporal pooling and unpooling operations proposed above, we develop a U-shaped multi-scale architecture, Spatio-Temporal U-Net, to address the challenge of analyzing and predicting graph-structured sequences. Following the classic U-shaped design, it contains two parts in symmetry: downsampling and upsampling. In the contracting part, it firstly applies graph convolution to aggregate information from each node's neighborhoods, and then follows by the ST-Pool layer to encode convolved features into multiple spatio-temporal resolution. In the expansive part, it utilizes the ST-Unpool layer for upsampling the reduced features to their original dimensions, with the concatenation of corresponding high-level features retrieved from the downsampling. In the end, one graph convolution layer is attached to propagate the information through multiple spatial scales for the final prediction. The illustration of proposed architecture presents in Figure \ref{fig:model}. We now can summarize the main characteristics of ST-UNet in three aspects,
\begin{itemize}
  \item To the best of our knowledge, it is the first time that a multi-scale network with U-shaped design is applied to learn and model spatio-temporal structures from graph-structured time series.
  \item A novel pair of operators in spatio-temporal pooling and unpooling are firstly proposed for extracting and fusing multilevel features in spacetime domain.
  \item The proposed framework ST-UNet achieves the balance between accuracy and efficiency with considerable scalability through multi-scale feature extraction and fusion as shown in the experiment below.
\end{itemize}

\section{Experimental Studies}
In this section, we present the evaluation of our model proposed in Section \ref{sec:arch}. Several mainstream models are tested and analyzed on spatio-temporal prediction tasks. Experiments show that ST-UNet consistently outperforms other models and achieves state-of-the-art performance regarding prediction accuracy. We also perform the ablation study to validate the effectiveness of spatio-temporal pooling and unpooling operations. Comparison between GCN-based models suggests that ST-UNet has the superiority in balancing efficiency and scalability on the large-scale dataset. For a fair comparison, we execute grid search strategy to determine the best hyper-parameters on validations for all test models.

\subsection{Spatio-temporal Sequence Modeling on Moving-MNIST}
\begin{figure}
  \centering
  \includegraphics[width=0.47\textwidth]{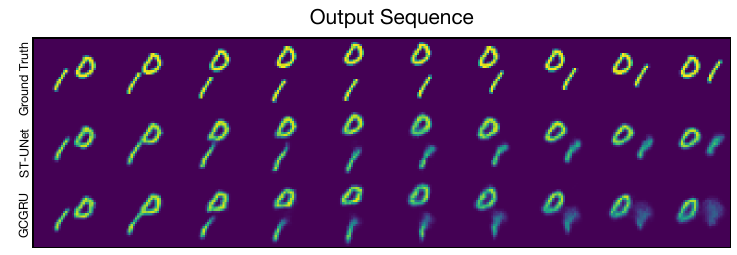}
  \caption{\label{fig:mmnist}Qualitative results for moving MNIST. First row is the ground truth, second and third are the predictions of ST-UNet($p=2, K=3$) and GCGRU($K=3$) respectively.}
\end{figure}

In order to investigate the ability of node-level prediction, we compare ST-UNet with its plain version GCGRU on a synthetic dataset, moving-MNIST constructed by \cite{xingjian2015convolutional}. It consists of 20-frame sequences (first 10 frames as input and the last for prediction), each of which contains two handwritten digits whose location is bouncing inside a 64 $\times$ 64 patch.\footnote{To make it feasible for all test models, the image frame in moving-MNIST is downsampled to 32 $\times$ 32 in the experiment of this section.} Following the default setup in \cite{seo2018structured}, image frames are converted into spatio-temporal graphs. The adjacency matrix is constructed based on distances between each pixel node and its equal neighbors of a k-nearest-neighbor graph in four directions (up, down, left and right). Kernel size of graph convolution $K$ is set to 3 for both models. The visualized outcome of moving sequence prediction in Figure \ref{fig:mmnist} indicates that, thanks to hierarchical feature fusion in spacetime domain, the U-shaped network can learn better representation and obtain superior performance than the model purely based on GCNs in the node-level. It suggests the transferability of such multi-scale designs from regular grids to non-Euclidean domain as well.

\subsection{Graph-structured Time-series Modeling on Traffic Prediction}
\paragraph{Experimental Setup} 
For traffic prediction task, we conduct experiments on two real-world public datasets: \textbf{METR-LA} released by \cite{li2018dcrnn_traffic} includes traffic information gathered by 207 loop detectors of Los Angeles County in 4 months, ranging from March 1st to June 30th of 2012; \textbf{PeMS (M/L)} generated by \cite{yu2018spatio} contains traffic status collected from monitoring stations deployed over California state highway system in the weekdays of May and June of 2012, containing 228 and 1026 stations respectively. Both datasets aggregate traffic records into a 5-min interval with an adjacency matrix describing the sensor topology of traffic networks. We use the same experimental settings of previous studies on these two datasets, including data preprocessing, dataset split, and other related configurations.

The following mainstream methods are selected as the baseline: 1). Historical Average (HA); 2). Linear Support Vector Regression (LSVR); 3). Auto-Regressive Integrated Moving Average (ARIMA); 4). Feedforward Neural Network (FNN); 5). Fully-Connected LSTM (FC-LSTM) \cite{sutskever2014sequence}; 6). Spatio-Temporal Graph Convolutional Networks (STGCN) \cite{yu2018spatio}; 7). Diffusion Convolutional Recurrent Neural Network (DCRNN) \cite{li2018dcrnn_traffic}.

This task requires using observed traffic time series in the window of one hour to forecast future status in the next 15, 30, and 60 minutes. Thus, three standard metrics of sequence prediction are adopted to measure the performance of different methods, namely, Mean Absolute Errors (MAE), Mean Absolute Percentage Errors (MAPE), and Root Mean Squared Errors (RMSE).

\paragraph{ST-UNet Settings}
All ST-UNet models use the kernel size $K=3$ for the graph convolution. Both spatial pooling level $p$ and temporal dilation $s$ are set at 2 with `direct copy' employed as the upsampling approach. We train our models by using Adam optimizer to minimize the mean of $L_1$ and $L_2$ loss for 80 epochs with the batch size as 50. The schedule sampling and layer normalization are utilized in training for better generalization. The initial learning rate is $10^{-2}$ with a decay rate of 0.7 after every 8 epochs. The hidden size of recurrent units in our model is 96 for METR-LA dataset; while it is assigned to 64 for the rest.

\paragraph{Results Analysis}
Table \ref{tab:res} demonstrates the numerical results of spatio-temporal traffic prediction on datasets METR-LA and PeMS-M. We observe the following phenomenon in both datasets: 1) graph convolution based models, including STGCN, DCRNN and ST-UNet generally outperform other baselines, which emphasizes the importance of including graph topology for traffic prediction. 2) RNN-based models tend to act better for the long-term prediction, suggesting their advantages in capturing temporal dependency. 3) regarding the adopted metrics, ST-UNet achieves the best performance for all three forecasting windows, which validates the effectiveness of multi-scale designs in spatio-temporal sequence modeling. 4) traditional approaches such as LSVR and ARIMA mostly perform worse than deep learning models, due to their limited capacities for handling complex non-linear data. In addition, historical average is a reflection of traffic status in a long-term, which is invariant to the short-term impact. 

\begin{table*}
\centering
\caption{\label{tab:res}Performance comparison of different models on METR-LA and PeMS-M datasets.}
\resizebox{0.92\textwidth}{!}{
\begin{tabular}{c||c|c|c||c|c|c}
  \hline \hline
  \multirow{2}{*}{Model} & \multicolumn{3}{|c||}{METR-LA (15/ 30/ 60 min)} & \multicolumn{3}{c}{PeMS-M (15/ 30/ 60 min)} \\ \cline{2-7}
  & MAE & MAPE (\%) & RMSE & MAE & MAPE (\%) & RMSE \\ \hline \hline
  HA & 4.16 & 13.0 & 7.80 & 4.01 & 10.61 & 7.20 \\ \hline
  LSVR & 2.97/ 3.64/ 4.67 & 7.68/ 9.9/ 13.63 & 5.89/ 7.35/ 9.13 & 2.50/ 3.63/ 4.54 & 5.81/ 8.88/ 11.50 & 4.55/ 6.67/ 8.28 \\ \hline
  ARIMA & 3.99/ 5.15/ 6.90 & 9.6/ 12.7/ 17.4 & 8.21/ 10.45/ 13.23 & 5.55/ 5.86/ 6.83 & 12.92/ 13.94/ 17.34 & 9.00/ 9.13/ 11.48 \\ \hline
  FNN & 3.99/ 4.23/ 4.49 & 9.9/ 12.9/ 14.0 & 7.94/ 8.17/ 8.69 & 2.39/ 3.41/ 4.88 & 5.53/ 8.16/ 12.12 & 4.40/ 6.40/ 8.84 \\ \hline
  FC-LSTM & 3.44/ 3.77/ 4.37 & 9.6/ 10.9/ 13.2 & 6.30/ 7.23/ 8.69 & 3.67/ 3.87/ 4.19 & 9.09/ 9.57/ 10.55 & 6.58/ 7.03/ 7.79 \\ \hline
  STGCN & 2.87/ 3.48/ 4.45 & 7.4/ 9.4/ 11.8 & 5.54/ 6.84/ 8.41 & 2.25/ 3.03/ 4.02 & 5.26/ 7.33/ 9.85 & 4.04/ 5.70/ 7.64 \\ \hline
  DCRNN & 2.77/ 3.15/ 3.60 & 7.3/ 8.8/ 10.5 & 5.38/ 6.45/ 7.59 & 2.25/ 2.98/ 3.83 & 5.30/ 7.39/ 9.85 & 4.04/ 5.58/ 7.19 \\ \hline 
  ST-UNet & \textbf{2.72}/ \textbf{3.12}/ \textbf{3.55} & \textbf{6.9}/ \textbf{8.4}/ \textbf{10.0} & \textbf{5.13}/ \textbf{6.16}/ \textbf{7.40} & \textbf{2.15}/ \textbf{2.81}/ \textbf{3.38} & \textbf{5.06}/ \textbf{6.79}/ \textbf{8.33} & \textbf{4.03}/ \textbf{5.42}/ \textbf{6.68} \\ \hline \hline
\end{tabular}}
\end{table*}

\subsection{Ablation Study of ST-Pool \& ST-Unpool}
As the above two tasks reveal, ST-UNet steadily outperforms mainstream models for spatio-temporal prediction. But it may be argued that performance gains are actually due to the deeper architecture or benefit from multilevel abstraction in spatial or temporal alone. Therefore, we initiate an ablation study to investigate the contribution of spatio-temporal pooling and unpooling operations in our model. We conduct the experiment with ST-UNet in four styles: the plain version by removing all ST-Pool and ST-Unpool operations; T-UNet only with pooling and upooling in temporal; S-UNet only with pooling and unpooling in spatial; and the full version. To the aim of pure comparison, we only test these variants without additional training tricks. The numerical outcome in Table \ref{tab:ablation} confirms that the proposed operations are valid for model enhancement in both spatial and temporal dimension. Moreover, thanks to the multi-scale feature integration through U-shaped network, applying pooling and unpooling operations in space and time coherently results in further improvement and better generalization. 

\begin{table}
\centering
\caption{\label{tab:ablation}Comparison of ST-UNet variants with or without ST-Pool \& ST-Unpool operations in terms of prediction accuracy on PeMS-M.}
\resizebox{0.46\textwidth}{!}{
\begin{tabular}{c|c||c|c|c|c}
  \hline \hline
  \multicolumn{2}{c||}{Models} & GCGRU & T-UNet &S-UNet& ST-UNet \\ \hline \hline
    \multirow{3}{*}{\rotatebox{90}{$15$ min}}
        & MAE  & $2.292\pm 0.007$ & $2.266\pm 0.005$ & $2.272\pm 0.004$ & \textbf{2.248$\pm$ 0.004} \\ \cline{2-6}
  & MAPE(\%) & $5.316\pm 0.031$ & $5.302\pm 0.019$& $5.300\pm 0.014$ & \textbf{5.244$\pm$ 0.028} \\ \cline{2-6}
  & RMSE & $4.086\pm 0.010$ & $4.014\pm 0.010$ & $4.000\pm 0.006$ & \textbf{3.994$\pm$ 0.005} \\ \hline \hline
  \multirow{3}{*}{\rotatebox{90}{$30$ min}}  
  & MAE  & $3.050\pm 0.037$ & $3.004\pm 0.010$ & $3.020\pm 0.017$ & \textbf{2.980$\pm$ 0.011} \\ \cline{2-6}
  & MAPE(\%) & $7.258\pm 0.031$ & $7.294\pm 0.046$ & $7.274\pm 0.050$ & \textbf{7.124$\pm$ 0.037} \\ \cline{2-6}
  & RMSE & $5.688\pm 0.044$ & $5.556\pm 0.051$ & $5.544\pm 0.010$ & \textbf{5.452$\pm$ 0.019} \\ \hline \hline
  \multirow{3}{*}{\rotatebox{90}{$60$ min}}
  & MAE  & $3.866\pm 0.094$ & $3.850\pm 0.061$ & $3.802\pm 0.052$ & \textbf{3.756$\pm$ 0.031} \\ \cline{2-6}
  & MAPE(\%) & $9.328\pm 0.098$ & $9.228\pm 0.089$ & $9.200\pm 0.148$ & \textbf{8.844$\pm$ 0.082} \\ \cline{2-6}
  & RMSE & $7.210\pm 0.108$ & $7.036\pm 0.149$ & $7.052\pm 0.129$ & \textbf{6.716$\pm$ 0.025} \\ \hline \hline
\end{tabular}}
\end{table}

\subsection{\label{sec:up}Comparison Study of Upsampling Approaches in ST-Unpool}
As we discussed in Section \ref{sec:un}, there are three methods for upsampling spatial features in the unpooling part. We carry out the experiment to examine the relation between these methods and the performance of corresponding models. Comparison of three upsampling approaches in terms of the mean square error is summarized in Table \ref{tab:ups}. The method of direct copy generally performs better than the other two, especially in relatively long terms. It suggests that the simple mechanism may be more steady and robust in this case. Furthermore, local properties within a super node such as degree orders and connectedness may not contain enough information to support complex feature reconstruction, due to the isomorphism of its node elements and significant structural differences among other nodes.

\begin{table}
\centering
\caption{\label{tab:ups}Comparison of different upsampling approaches in ST-Unpool in terms of MSE on PeMS-M (The notion `${\S}$' indicates that the test model may not converge eventually).}
\resizebox{0.46\textwidth}{!}{
\begin{tabular}{c||c|c|c}
  \hline \hline
  Models & Direct-Copy & Ordered-Deconv & Weighted-Deconv\\ \hline \hline
  15min & $3.994\pm 0.005$ & \textbf{3.980$\pm$ 0.009} & $4.603\pm 0.074^{\S}$\\ \hline
  30min & \textbf{5.452$\pm$ 0.019} & $5.468\pm 0.015$ & $5.658\pm 0.192^{\S}$\\ \hline
  60min & \textbf{6.716$\pm$ 0.025} & $6.956\pm 0.074$ & $7.425\pm 0.488^{\S}$\\ \hline
\end{tabular}}
\end{table}

\subsection{Scalability and Efficiency Study on Large-scale Graph Data}
To test the scalability and efficiency of ST-UNet, we experiment our model and other GCN-based ones on a large dataset PeMS-L which contains over one thousand sensor nodes in a single graph. We list the comparison of prediction accuracy for four major models in Table \ref{tab:large}. Apparently, conventional graph convolution based approaches, including GCGRU and DCRNN face great challenges in handling such large-scale graphs. We use the symbol `$\sharp$' to mark the model whose batch size is forced to reduce a half since its graphical memory consumption exceeded the capacity over a standard GPU card.\footnote{All experiments are compiled and tested on a CentOS cluster (CPU: Intel(R) Xeon(R) CPU E5-2620 v4 @ 2.10GHz, GPU: NVIDIA GeForce GTX 1080).} By means of its fully convolutional structures, STGCN is able to process such large dataset at once. With the help of exploring spatio-temporal correlations in a global view, it behaves well in short-and-mid term prediction but suffering from overfitting in long periods. On the other hand, DCRNN maintains a higher standard on long-term forecasting but with the cost of massive computational demands. For instance, the model normally takes more than 10 minutes to train one epoch with the batch size of 16 on PeMS-L. By contrast, ST-UNet confers better outcome in less half of the time that DCRNN need. It has reached the balance between time efficiency and prediction accuracy through spatial and temporal pooling operations applied. It also has advantages in extracting spatial features and temporal dependencies with fewer parameters and in multilevel abstraction.

\begin{table}
\centering
\caption{\label{tab:large}Comparison of GCN-based models in terms of prediction accuracy on the large-scale dataset PeMS-L.}
\resizebox{0.46\textwidth}{!}{
\begin{tabular}{c||c|c|c}
  \hline \hline
  \multirow{2}{*}{Models} & \multicolumn{3}{c}{PeMS-L (15/ 30/ 60 min)} \\ \cline{2-4}
  & MAE & MAPE (\%) & RMSE \\ \hline \hline
  HA & 4.60 & 12.50 & 8.05 \\ \hline
  $\text{GCGRU}^{\sharp}$ & 2.48/ 3.43/ 4.08 & 5.76/ 8.45/ 10.28 & 4.40/ 6.25/ 7.62 \\ \hline
  STGCN & 2.37/ 3.27/ 4.36 & 5.56/ 7.98/ 11.59 & 4.32/ 6.21/ 8.31 \\ \hline
  $\text{DCRNN}^{\sharp}$ & 2.41/ 3.28/ 4.32 & 5.61/ 8.18/ 11.33 & \textbf{4.22}/ 5.87/ 7.58 \\ \hline 
  ST-UNet & \textbf{2.34}/ \textbf{3.02}/ \textbf{3.66} & \textbf{5.54}/ \textbf{7.56}/ \textbf{9.52} & 4.32/ \textbf{5.81}/ \textbf{7.14} \\ \hline \hline
\end{tabular}}
\end{table}

\section{Conclusion}
In this paper, we propose a universal multi-scale architecture ST-UNet to learn and predict graph-structured time series, integrating multi-granularity graph convolution and dilated recurrent skip-connections through the U-shaped network design. Experiments show that our model consistently outperforms other state-of-the-art methods on several real-world datasets, indicating its great potentials on extracting comprehensive spatio-temporal features through scale-spanning sequence modeling. The ablation study validates the efficiency improvement obtained from the proposed pooling and unpooling operations in spacetime domain. Moreover, ST-UNet also achieves the balance between efficiency and capacity with considerable flexibility. These features are quite promising and practical for structured sequence modeling in the future research development and industrial applications.

%
\bibliographystyle{ACM-Reference-Format}
\bibliography{ref_acm}


\begin{thebibliography}{26}


\ifx \showCODEN    \undefined \def \showCODEN     #1{\unskip}     \fi
\ifx \showDOI      \undefined \def \showDOI       #1{#1}\fi
\ifx \showISBNx    \undefined \def \showISBNx     #1{\unskip}     \fi
\ifx \showISBNxiii \undefined \def \showISBNxiii  #1{\unskip}     \fi
\ifx \showISSN     \undefined \def \showISSN      #1{\unskip}     \fi
\ifx \showLCCN     \undefined \def \showLCCN      #1{\unskip}     \fi
\ifx \shownote     \undefined \def \shownote      #1{#1}          \fi
\ifx \showarticletitle \undefined \def \showarticletitle #1{#1}   \fi
\ifx \showURL      \undefined \def \showURL       {\relax}        \fi
\providecommand\bibfield[2]{#2}
\providecommand\bibinfo[2]{#2}
\providecommand\natexlab[1]{#1}
\providecommand\showeprint[2][]{arXiv:#2}

\bibitem[\protect\citeauthoryear{Bruna, Zaremba, Szlam, and LeCun}{Bruna
  et~al\mbox{.}}{2013}]%
        {bruna2013spectral}
\bibfield{author}{\bibinfo{person}{Joan Bruna}, \bibinfo{person}{Wojciech
  Zaremba}, \bibinfo{person}{Arthur Szlam}, {and} \bibinfo{person}{Yann
  LeCun}.} \bibinfo{year}{2013}\natexlab{}.
\newblock \showarticletitle{Spectral networks and locally connected networks on
  graphs}.
\newblock \bibinfo{journal}{\emph{arXiv preprint arXiv:1312.6203}}
  (\bibinfo{year}{2013}).
\newblock


\bibitem[\protect\citeauthoryear{Chang, Zhang, Han, Yu, Guo, Tan, Cui,
  Witbrock, Hasegawa-Johnson, and Huang}{Chang et~al\mbox{.}}{2017}]%
        {chang2017dilated}
\bibfield{author}{\bibinfo{person}{Shiyu Chang}, \bibinfo{person}{Yang Zhang},
  \bibinfo{person}{Wei Han}, \bibinfo{person}{Mo Yu}, \bibinfo{person}{Xiaoxiao
  Guo}, \bibinfo{person}{Wei Tan}, \bibinfo{person}{Xiaodong Cui},
  \bibinfo{person}{Michael Witbrock}, \bibinfo{person}{Mark~A
  Hasegawa-Johnson}, {and} \bibinfo{person}{Thomas~S Huang}.}
  \bibinfo{year}{2017}\natexlab{}.
\newblock \showarticletitle{Dilated recurrent neural networks}. In
  \bibinfo{booktitle}{\emph{Advances in Neural Information Processing
  Systems}}. \bibinfo{pages}{77--87}.
\newblock


\bibitem[\protect\citeauthoryear{Chung, Gulcehre, Cho, and Bengio}{Chung
  et~al\mbox{.}}{2014}]%
        {chung2014empirical}
\bibfield{author}{\bibinfo{person}{Junyoung Chung}, \bibinfo{person}{Caglar
  Gulcehre}, \bibinfo{person}{KyungHyun Cho}, {and} \bibinfo{person}{Yoshua
  Bengio}.} \bibinfo{year}{2014}\natexlab{}.
\newblock \showarticletitle{Empirical evaluation of gated recurrent neural
  networks on sequence modeling}.
\newblock \bibinfo{journal}{\emph{arXiv preprint arXiv:1412.3555}}
  (\bibinfo{year}{2014}).
\newblock


\bibitem[\protect\citeauthoryear{Defferrard, Bresson, and
  Vandergheynst}{Defferrard et~al\mbox{.}}{2016}]%
        {defferrard2016convolutional}
\bibfield{author}{\bibinfo{person}{Micha{\"e}l Defferrard},
  \bibinfo{person}{Xavier Bresson}, {and} \bibinfo{person}{Pierre
  Vandergheynst}.} \bibinfo{year}{2016}\natexlab{}.
\newblock \showarticletitle{Convolutional neural networks on graphs with fast
  localized spectral filtering}. In \bibinfo{booktitle}{\emph{Advances in
  Neural Information Processing Systems}}. \bibinfo{pages}{3844--3852}.
\newblock


\bibitem[\protect\citeauthoryear{Gao and Ji}{Gao and Ji}{2019}]%
        {gao2019graph}
\bibfield{author}{\bibinfo{person}{Hongyang Gao} {and}
  \bibinfo{person}{Shuiwang Ji}.} \bibinfo{year}{2019}\natexlab{}.
\newblock \bibinfo{title}{Graph U-Net}.
\newblock
\newblock
\urldef\tempurl%
\url{https://openreview.net/forum?id=HJePRoAct7}
\showURL{%
\tempurl}


\bibitem[\protect\citeauthoryear{Gao, Wang, and Ji}{Gao et~al\mbox{.}}{2018}]%
        {gao2018large}
\bibfield{author}{\bibinfo{person}{Hongyang Gao}, \bibinfo{person}{Zhengyang
  Wang}, {and} \bibinfo{person}{Shuiwang Ji}.} \bibinfo{year}{2018}\natexlab{}.
\newblock \showarticletitle{Large-scale learnable graph convolutional
  networks}. In \bibinfo{booktitle}{\emph{Proceedings of the 24th ACM SIGKDD
  International Conference on Knowledge Discovery \& Data Mining}}. ACM,
  \bibinfo{pages}{1416--1424}.
\newblock


\bibitem[\protect\citeauthoryear{Hamilton, Ying, and Leskovec}{Hamilton
  et~al\mbox{.}}{2017}]%
        {hamilton2017inductive}
\bibfield{author}{\bibinfo{person}{Will Hamilton}, \bibinfo{person}{Zhitao
  Ying}, {and} \bibinfo{person}{Jure Leskovec}.}
  \bibinfo{year}{2017}\natexlab{}.
\newblock \showarticletitle{Inductive representation learning on large graphs}.
  In \bibinfo{booktitle}{\emph{Advances in Neural Information Processing
  Systems}}. \bibinfo{pages}{1024--1034}.
\newblock


\bibitem[\protect\citeauthoryear{Hammond, Vandergheynst, and Gribonval}{Hammond
  et~al\mbox{.}}{2011}]%
        {hammond2011wavelets}
\bibfield{author}{\bibinfo{person}{David~K Hammond}, \bibinfo{person}{Pierre
  Vandergheynst}, {and} \bibinfo{person}{R{\'e}mi Gribonval}.}
  \bibinfo{year}{2011}\natexlab{}.
\newblock \showarticletitle{Wavelets on graphs via spectral graph theory}.
\newblock \bibinfo{journal}{\emph{Applied and Computational Harmonic Analysis}}
  \bibinfo{volume}{30}, \bibinfo{number}{2} (\bibinfo{year}{2011}),
  \bibinfo{pages}{129--150}.
\newblock


\bibitem[\protect\citeauthoryear{Hochreiter and Schmidhuber}{Hochreiter and
  Schmidhuber}{1997}]%
        {hochreiter1997long}
\bibfield{author}{\bibinfo{person}{Sepp Hochreiter} {and}
  \bibinfo{person}{J{\"u}rgen Schmidhuber}.} \bibinfo{year}{1997}\natexlab{}.
\newblock \showarticletitle{Long short-term memory}.
\newblock \bibinfo{journal}{\emph{Neural Computation}} \bibinfo{volume}{9},
  \bibinfo{number}{8} (\bibinfo{year}{1997}), \bibinfo{pages}{1735--1780}.
\newblock


\bibitem[\protect\citeauthoryear{Jain, Zamir, Savarese, and Saxena}{Jain
  et~al\mbox{.}}{2016}]%
        {jain2016structural}
\bibfield{author}{\bibinfo{person}{Ashesh Jain}, \bibinfo{person}{Amir~R
  Zamir}, \bibinfo{person}{Silvio Savarese}, {and} \bibinfo{person}{Ashutosh
  Saxena}.} \bibinfo{year}{2016}\natexlab{}.
\newblock \showarticletitle{Structural-RNN: Deep learning on spatio-temporal
  graphs}. In \bibinfo{booktitle}{\emph{Proceedings of the IEEE Conference on
  Computer Vision and Pattern Recognition}}. \bibinfo{pages}{5308--5317}.
\newblock


\bibitem[\protect\citeauthoryear{Kipf and Welling}{Kipf and Welling}{2016}]%
        {kipf2016semi}
\bibfield{author}{\bibinfo{person}{Thomas~N Kipf} {and} \bibinfo{person}{Max
  Welling}.} \bibinfo{year}{2016}\natexlab{}.
\newblock \showarticletitle{Semi-supervised classification with graph
  convolutional networks}.
\newblock \bibinfo{journal}{\emph{arXiv preprint arXiv:1609.02907}}
  (\bibinfo{year}{2016}).
\newblock


\bibitem[\protect\citeauthoryear{Koutnik, Greff, Gomez, and
  Schmidhuber}{Koutnik et~al\mbox{.}}{2014}]%
        {koutnik2014clockwork}
\bibfield{author}{\bibinfo{person}{Jan Koutnik}, \bibinfo{person}{Klaus Greff},
  \bibinfo{person}{Faustino Gomez}, {and} \bibinfo{person}{Juergen
  Schmidhuber}.} \bibinfo{year}{2014}\natexlab{}.
\newblock \showarticletitle{A clockwork rnn}.
\newblock \bibinfo{journal}{\emph{arXiv preprint arXiv:1402.3511}}
  (\bibinfo{year}{2014}).
\newblock


\bibitem[\protect\citeauthoryear{Li, Cui, Zheng, Xu, and Yang}{Li
  et~al\mbox{.}}{2018a}]%
        {AAAI1817103}
\bibfield{author}{\bibinfo{person}{Chaolong Li}, \bibinfo{person}{Zhen Cui},
  \bibinfo{person}{Wenming Zheng}, \bibinfo{person}{Chunyan Xu}, {and}
  \bibinfo{person}{Jian Yang}.} \bibinfo{year}{2018}\natexlab{a}.
\newblock \showarticletitle{Spatio-Temporal graph convolution for skeleton
  based action recognition}. In \bibinfo{booktitle}{\emph{AAAI Conference on
  Artificial Intelligence}}.
\newblock


\bibitem[\protect\citeauthoryear{Li, Yu, Shahabi, and Liu}{Li
  et~al\mbox{.}}{2018b}]%
        {li2018dcrnn_traffic}
\bibfield{author}{\bibinfo{person}{Yaguang Li}, \bibinfo{person}{Rose Yu},
  \bibinfo{person}{Cyrus Shahabi}, {and} \bibinfo{person}{Yan Liu}.}
  \bibinfo{year}{2018}\natexlab{b}.
\newblock \showarticletitle{Diffusion convolutional recurrent neural network:
  Data-driven traffic forecasting}. In \bibinfo{booktitle}{\emph{International
  Conference on Learning Representations}}.
\newblock


\bibitem[\protect\citeauthoryear{Maue and Sanders}{Maue and Sanders}{2007}]%
        {maue2007engineering}
\bibfield{author}{\bibinfo{person}{Jens Maue} {and} \bibinfo{person}{Peter
  Sanders}.} \bibinfo{year}{2007}\natexlab{}.
\newblock \showarticletitle{Engineering algorithms for approximate weighted
  matching}. In \bibinfo{booktitle}{\emph{International Workshop on
  Experimental and Efficient Algorithms}}. Springer, \bibinfo{pages}{242--255}.
\newblock


\bibitem[\protect\citeauthoryear{Neil, Pfeiffer, and Liu}{Neil
  et~al\mbox{.}}{2016}]%
        {neil2016phased}
\bibfield{author}{\bibinfo{person}{Daniel Neil}, \bibinfo{person}{Michael
  Pfeiffer}, {and} \bibinfo{person}{Shih-Chii Liu}.}
  \bibinfo{year}{2016}\natexlab{}.
\newblock \showarticletitle{Phased lstm: Accelerating recurrent network
  training for long or event-based sequences}. In
  \bibinfo{booktitle}{\emph{Advances in Neural Information Processing
  Systems}}. \bibinfo{pages}{3882--3890}.
\newblock


\bibitem[\protect\citeauthoryear{Niepert, Ahmed, and Kutzkov}{Niepert
  et~al\mbox{.}}{2016}]%
        {niepert2016learning}
\bibfield{author}{\bibinfo{person}{Mathias Niepert}, \bibinfo{person}{Mohamed
  Ahmed}, {and} \bibinfo{person}{Konstantin Kutzkov}.}
  \bibinfo{year}{2016}\natexlab{}.
\newblock \showarticletitle{Learning convolutional neural networks for graphs}.
  In \bibinfo{booktitle}{\emph{International conference on Machine Learning}}.
  \bibinfo{pages}{2014--2023}.
\newblock


\bibitem[\protect\citeauthoryear{Ronneberger, Fischer, and Brox}{Ronneberger
  et~al\mbox{.}}{2015}]%
        {ronneberger2015u}
\bibfield{author}{\bibinfo{person}{Olaf Ronneberger}, \bibinfo{person}{Philipp
  Fischer}, {and} \bibinfo{person}{Thomas Brox}.}
  \bibinfo{year}{2015}\natexlab{}.
\newblock \showarticletitle{U-net: Convolutional networks for biomedical image
  segmentation}. In \bibinfo{booktitle}{\emph{International Conference on
  Medical Image Computing and Computer-assisted Intervention}}. Springer,
  \bibinfo{pages}{234--241}.
\newblock


\bibitem[\protect\citeauthoryear{Seo, Defferrard, Vandergheynst, and
  Bresson}{Seo et~al\mbox{.}}{2016}]%
        {seo2018structured}
\bibfield{author}{\bibinfo{person}{Youngjoo Seo}, \bibinfo{person}{Micha{\"e}l
  Defferrard}, \bibinfo{person}{Pierre Vandergheynst}, {and}
  \bibinfo{person}{Xavier Bresson}.} \bibinfo{year}{2016}\natexlab{}.
\newblock \showarticletitle{Structured sequence modeling with graph
  convolutional recurrent networks}.
\newblock \bibinfo{journal}{\emph{arXiv preprint arXiv:1612.07659}}.
\newblock


\bibitem[\protect\citeauthoryear{Shuman, Narang, Frossard, Ortega, and
  Vandergheynst}{Shuman et~al\mbox{.}}{2012}]%
        {shuman2012emerging}
\bibfield{author}{\bibinfo{person}{David~I Shuman}, \bibinfo{person}{Sunil~K
  Narang}, \bibinfo{person}{Pascal Frossard}, \bibinfo{person}{Antonio Ortega},
  {and} \bibinfo{person}{Pierre Vandergheynst}.}
  \bibinfo{year}{2012}\natexlab{}.
\newblock \showarticletitle{The emerging field of signal processing on graphs:
  Extending high-dimensional data analysis to networks and other irregular
  domains}.
\newblock \bibinfo{journal}{\emph{arXiv preprint arXiv:1211.0053}}
  (\bibinfo{year}{2012}).
\newblock


\bibitem[\protect\citeauthoryear{Sutskever, Vinyals, and Le}{Sutskever
  et~al\mbox{.}}{2014}]%
        {sutskever2014sequence}
\bibfield{author}{\bibinfo{person}{Ilya Sutskever}, \bibinfo{person}{Oriol
  Vinyals}, {and} \bibinfo{person}{Quoc~V Le}.}
  \bibinfo{year}{2014}\natexlab{}.
\newblock \showarticletitle{Sequence to sequence learning with neural
  networks}. In \bibinfo{booktitle}{\emph{Advances in Neural Information
  Processing Systems}}. \bibinfo{pages}{3104--3112}.
\newblock


\bibitem[\protect\citeauthoryear{Xingjian, Chen, Wang, Yeung, Wong, and
  Woo}{Xingjian et~al\mbox{.}}{2015}]%
        {xingjian2015convolutional}
\bibfield{author}{\bibinfo{person}{SHI Xingjian}, \bibinfo{person}{Zhourong
  Chen}, \bibinfo{person}{Hao Wang}, \bibinfo{person}{Dit-Yan Yeung},
  \bibinfo{person}{Wai-Kin Wong}, {and} \bibinfo{person}{Wang-chun Woo}.}
  \bibinfo{year}{2015}\natexlab{}.
\newblock \showarticletitle{Convolutional LSTM network: A machine learning
  approach for precipitation nowcasting}. In \bibinfo{booktitle}{\emph{Advances
  in Neural Information Processing Systems}}. \bibinfo{pages}{802--810}.
\newblock


\bibitem[\protect\citeauthoryear{Yu, Yin, and Zhu}{Yu et~al\mbox{.}}{2018}]%
        {yu2018spatio}
\bibfield{author}{\bibinfo{person}{Bing Yu}, \bibinfo{person}{Haoteng Yin},
  {and} \bibinfo{person}{Zhanxing Zhu}.} \bibinfo{year}{2018}\natexlab{}.
\newblock \showarticletitle{Spatio-temporal graph convolutional networks: A
  deep learning framework for traffic forecasting}. In
  \bibinfo{booktitle}{\emph{Proceedings of the 27th International Joint
  Conference on Artificial Intelligence}}. \bibinfo{pages}{3634--3640}.
\newblock


\bibitem[\protect\citeauthoryear{Zeiler and Fergus}{Zeiler and Fergus}{2014}]%
        {zeiler2014visualizing}
\bibfield{author}{\bibinfo{person}{Matthew~D Zeiler} {and} \bibinfo{person}{Rob
  Fergus}.} \bibinfo{year}{2014}\natexlab{}.
\newblock \showarticletitle{Visualizing and understanding convolutional
  networks}. In \bibinfo{booktitle}{\emph{European Conference on Computer
  Vision}}. Springer, \bibinfo{pages}{818--833}.
\newblock


\bibitem[\protect\citeauthoryear{Zeiler, Taylor, Fergus, et~al\mbox{.}}{Zeiler
  et~al\mbox{.}}{2011}]%
        {zeiler2011adaptive}
\bibfield{author}{\bibinfo{person}{Matthew~D Zeiler}, \bibinfo{person}{Graham~W
  Taylor}, \bibinfo{person}{Rob Fergus}, {et~al\mbox{.}}}
  \bibinfo{year}{2011}\natexlab{}.
\newblock \showarticletitle{Adaptive deconvolutional networks for mid and high
  level feature learning}. In \bibinfo{booktitle}{\emph{International
  Conference on Computer Vision}}, Vol.~\bibinfo{volume}{1}.
  \bibinfo{pages}{6}.
\newblock


\bibitem[\protect\citeauthoryear{Zhang, Zheng, Qi, Li, Yi, and Li}{Zhang
  et~al\mbox{.}}{2018}]%
        {zhang2018predicting}
\bibfield{author}{\bibinfo{person}{Junbo Zhang}, \bibinfo{person}{Yu Zheng},
  \bibinfo{person}{Dekang Qi}, \bibinfo{person}{Ruiyuan Li},
  \bibinfo{person}{Xiuwen Yi}, {and} \bibinfo{person}{Tianrui Li}.}
  \bibinfo{year}{2018}\natexlab{}.
\newblock \showarticletitle{Predicting citywide crowd flows using deep
  spatio-temporal residual networks}.
\newblock \bibinfo{journal}{\emph{Artificial Intelligence}}
  \bibinfo{volume}{259} (\bibinfo{year}{2018}), \bibinfo{pages}{147--166}.
\newblock


\end{thebibliography}

%

\end{document}